\documentclass[letterpaper, 10 pt, conference]{ieeeconf}
\IEEEoverridecommandlockouts

\usepackage{cite}
\usepackage{amsmath,amssymb,amsfonts}
\usepackage{algorithmic}
\usepackage{graphicx}
\usepackage{textcomp}
\usepackage{xcolor}
\usepackage{import,bm}
\newcommand{\mt}[1]{\bm{#1}}

\def\BibTeX{{\rm B\kern-.05em{\sc i\kern-.025em b}\kern-.08em
    T\kern-.1667em\lower.7ex\hbox{E}\kern-.125emX}}
\begin{document}

\title{\LARGE \bf
Deep-Learning Control of Lower-Limb Exoskeletons via simplified Therapist Input\\
}

\author{Lorenzo Vianello$^{1}$, Clément Lhoste$^{2}$, Emek Barış Küçüktabak$^{7}$, Matthew R. Short$^{1,4}$, \\Levi Hargrove$^{5,6}$, Jose L. Pons$^{1,3,4,6}$%
\thanks{$^1$ Legs and Walking Lab of Shirley Ryan AbilityLab, Chicago, IL, USA}%
\thanks{$^2$ Rehabilitation Engineering Laboratory, ETH Zürich, CH}%
\thanks{$^3$ Center for Robotics and Biosystems of Northwestern University, Evanston, IL, USA}%
\thanks{$^4$ Department of Biomedical Engineering, Northwestern University, Evanston, IL, USA}%
\thanks{$^5$ Neural Engineering for Prosthetics and Orthotics Lab of Shirley Ryan AbilityLab, Chicago, IL, USA}
\thanks{$^6$ Department of Physical Medicine and Rehabilitation, Northwestern University, Chicago, IL, USA}
\thanks{$^7$ Honda Research Institute, San Jose, CA (USA)}
\thanks{This work was conducted when E.B. Küçüktabak and C. Lhoste were with the Northwestern University and Shirley Ryan AbilityLab.}
}

\maketitle
\thispagestyle{empty}
\pagestyle{empty}

\begin{abstract}
Partial-assistance exoskeletons hold significant potential for gait rehabilitation by promoting active participation during (re)learning of ``normal'' walking patterns. Typically, the control of interaction torques in partial-assistance exoskeletons relies on a hierarchical control structure. These approaches require extensive calibration due to the complexity of the controller and user-specific parameter tuning, especially for activities like stair or ramp navigation.
To address the limitations of hierarchical control in exoskeletons, this work proposes a three-step, data-driven approach: (1) using recent sensor data to probabilistically infer locomotion states (landing step length, landing step height, walking velocity, step clearance, gait phase), (2) allowing therapists to modify these features via a user interface, and (3) using the adjusted locomotion features to predict the desired joint posture and model stiffness in a spring-damper system based on prediction uncertainty.
We evaluated the proposed approach with two healthy participants performing treadmill walking and stair ascent/descent at varying speeds, with and without external modification of the gait features through a user interface.
Results showed a variation in kinematics according to the gait characteristics and a negative interaction power (-2.1 $\pm$ 1.6W for the hip and -0.6 $\pm$ 1.4W for the knee joints) suggesting exoskeleton assistance across the different conditions.  

\end{abstract}

\section{Introduction}
Lower-limb exoskeletons are valuable tools in rehabilitation for individuals with gait impairments. Clinically, these exoskeletons can be classified into two categories: full-assistance and partial-assistance~\cite{baud2021review}. Full-assistance exoskeletons are intended for patients with significant motor impairments who cannot ambulate independently, providing fully autonomous leg movement without requiring user input. They are usually controlled using position control strategies \cite{vouga2017twiice}. In contrast, partial-assistance exoskeletons are more appropriate for patients with mild to moderate impairments, as they support movement while still requiring the patient’s active participation \cite{tagliamonte2013human}.

This paper focuses on partial-assistance exoskeletons, which have greater potential in rehabilitation settings as they consider the patient's volitional movements in the control loop. These exoskeletons use approaches like haptic guidance~\cite{deMiguelFernndez2023} or resistance through error augmentation~\cite{9513580} to promote the (re)learning of walking patterns, depending on the patient's functional ability.

Both of these strategies, assistance and resistance, depend significantly on the control of interaction torques between the user and the exoskeleton. This is typically achieved through a hierarchical structure comprising high-, mid-, and low-level controls~\cite{Kim2022}. High-level control determines desired interaction torques for specific activities, such as walking on flat surfaces, stairs, or ramps. Mid-level control estimates various states within an activity, like swing and stance phases during walking, and adjusts the desired interaction torque accordingly \cite{lhoste2024deep}. Finally, low-level control compensates for the exoskeleton's dynamics and generates motor commands based on the profiles selected in high- and mid-level control.

To provide patients partial-assistance during ambulatory activities, desired interaction torques are typically rendered using virtual springs and dampers, characterized by stiffness and damping parameters, which are attached to reference joint positions. This necessitates the identification of many control parameters specific to each activity and phase of locomotion. As many of these parameters are also user-dependent, this can result in lengthy calibration periods for a single session of therapy. This issue has been well-documented for active prosthetic devices~\cite{simon2014configuring}, but few studies have quantified the time spent calibrating lower-limb exoskeleton controllers. This is likely due to a reliance on predefined trajectories for generating exoskeleton assistance profiles, which reduces the time and complexity of parameter selection, but limits the degree of parameter customization for individual users. Moreover, most research has concentrated on overground walking, with minimal investigation into scenarios involving stairs and ramps \cite{baud2021review}.

\begin{figure*}[t]
    \centering
    \vspace{0.2cm}
    \includegraphics[width=0.99\linewidth]{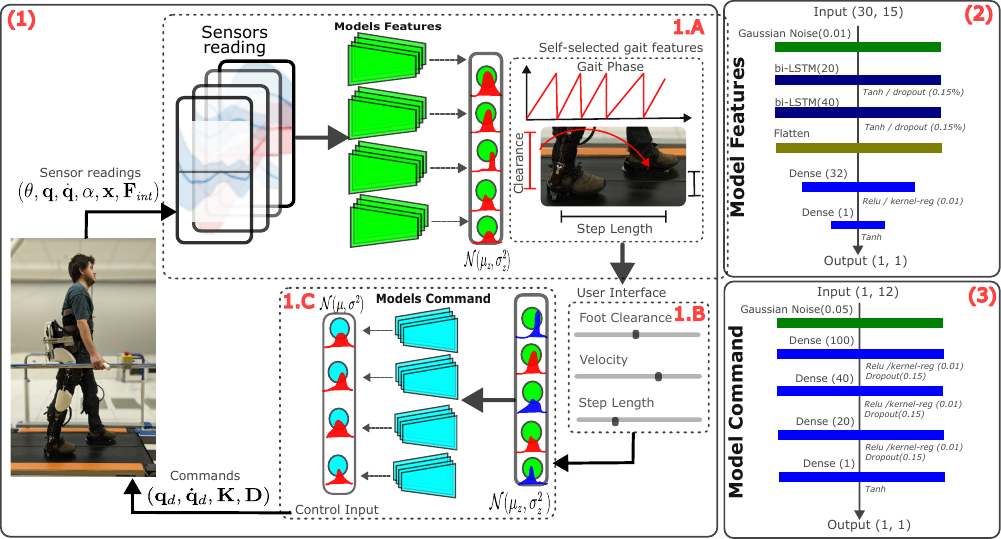}
    \caption{\small (1) Data-driven controller:(1.A) Sensor readings are passed to multiple deep-learning models to estimate representative features of the walking pattern. Thes resulting features are fitted to a normal distribution to consider the uncertanty of the locomotion pattern; (1.B) User interface allows to modify the self-selected locomotion features to allow the therapist to adapt the locomotion depending on the needs of each patient; (1.C) The resulting modified features are passed to a combination of models to regress the desired joint configuration of the exoskeleton. At the same time, the uncertainty of the prediction is propagated and used to model the stiffness of the impedance controller of the exoskeleton. Impedance parameters are passed to the low-level controller of the exoskeleton as commanded input. (2) Structure of the feature extractors models. (3) Structure of the command predictor models.  } 
    \label{fig:concept}
    \vspace{-0.5cm}
\end{figure*}

One potential solution to reduce controller complexity and calibration time is to employ data-driven methods that relate sensor readings with command outputs \cite{Huang2022, best2023data, molinaro2024task}. However, these approaches offer limited flexibility for personalization and adjustment of desired behaviors. For example, a therapist may require a progressive increase in foot clearance for a patient during a session of training. Additionally, therapists do not typically conceptualize adjustments in terms of gait cycle kinematics (e.g., trajectories or stiffness/damping parameters). Instead, they focus on clinically relevant features that are easier to interpret, such as step length, time, speed, or height.

For this reason, in this work, we propose a three-step data-driven approach to substitute the hierarchical structure of an exoskeleton controller (Fig. \ref{fig:concept}). 
In step (1) we probabilistically regress a short-term history of sensor readings (encoders, force sensors, foot-plate sensors) with a minimal set of clinically relevant gait features, namely step length, and height (to distinguish across ascending/descending surfaces), walking velocity, step clearance, gait phase; in step (2) we allow an operator (e.g., a physical therapist) to modify the self-selected gait features using a user interface and, finally in step (3) we probabilistically regress the current locomotion features to the desired joint posture of the exoskeleton, using prediction uncertainty to model the impedance in a spring-damper system connecting the user and exoskeleton.

\section{METHODS}

In this section, we present the structure of the proposed approach. In Sec.\ref{sec:method_exoskeleton}, we present the hardware used. In Sec. \ref{sec:data_collection}, we present how we collected data for training the deep-learning models. In Sec. \ref{sec:machine-learningModel}, we define the structure of the two consecutive models implemented to regress sensor readings and gait features and, subsequently, gait features with command inputs. Finally, Sec. \ref{sec:real-timeInference} presents the real-time use of the proposed approach and the walking tests performed for validation. 

\subsection{Exoskeleton}
\label{sec:method_exoskeleton}

We validated the proposed approach on a commercially available lower-limb exoskeleton (ExoMotus-X2, Fourier Intelligence, Singapore). The exoskeleton system has four active degrees of freedom (DoF) bilaterally at the hip and knee, and two passive DoFs at the ankle (Fig. \ref{fig:concept}). The following sensors are used to monitor both the exoskeleton's state and its interaction with the user~\cite{kuccuktabak2024haptic}: joint encoders measure the positions of the knee and hip angles ($\theta$) while strain gauges estimate interaction torque ($\tau_{int}$) at each joint. An Inertial Measurement Unit (IMU) on the backpack measures the orientation of the trunk ($\theta_{bp}$), while force-sensitive resistor (FSR) footplates measure ground reaction forces to quantify the weight distribution of the user ($\alpha$).
Communication between motors and sensors is achieved using CAN protocol. The controller is implemented on a ROS and C++ based open-source platform called the CANOpen Robot Controller (CORC) \cite{fong2022CANOPEN}; motor torque commands and sensor readings were updated at 333~Hz.
The exoskeleton was controlled using an impedance controller that models the desired interaction torque passed to the exoskeleton internal controller, described in~\cite{kuccuktabak2024haptic}:
\begin{equation}
    \tau_\text{int}^* = K(\mt{\theta} - \mt{\theta}_{d}) + D (\dot{\mt{\theta}} - \dot{\mt{\theta}}_{d}),
    \label{eq:impedanceController}
\end{equation}
where $K, D, \mt{\theta}_{d}, \dot{\mt{\theta}}_{d}$ are respectively the stiffness, damping, and the reference position and velocities returned by the data-driven controller as discussed in Sec. \ref{sec:real-timeInference}.

\subsection{Data collection and processing}
\label{sec:data_collection}

We collected data from nine healthy participants 
(5 males, 4 females, age 29.8$\pm$6.8, height 1.74$\pm$0.08 m, weight 70$\pm$10.6 kg) 
to train the deep-learning model. 
The Institutional Review Board of Northwestern University approved this study (STU00212684), and all procedures followed the Declaration of Helsinki. Participants performed various ambulatory activities while wearing the exoskeleton in two different controller conditions: haptic transparency (i.e., no assistance) \cite{kuccuktabak2024haptic} and state machine approach (i.e., non-zero assistance). In the state machine approach, the high-level controller was modeled using a remote (to switch between walking, ramps, or stairs). In the middle-level controller, the locomotion activities were split into four phases (early-stance, late-stance, early-swing, late-swing). For each of these phases, an equilibrium angle was selected and passed to an impedance controller.  
For each controller condition, participants performed five repetitions of overground walking (6 meters), three repetitions of stairs and ramps, both ascending and descending.

Data was processed offline to extract the \textit{gait-phase} ($\mt{gp}$) for both legs, a continuous value from $[0, 100]$ indicating progression between consecutive heel strikes. To consider the periodicity of the gait phase, we transformed this value to its polar representation ($\mt{c}^{gp}= \text{cos}(2 \pi \frac{\mt{gp}}{100}), \mt{s}^{gp}= \text{sin}(2 \pi \frac{\mt{gp}}{100}))$ \cite{kim2023gait}.
The horizontal and vertical foot position relative to the stance foot ($x_k, z_k | k\in [0,100]$) was calculated via forward kinematics and used to estimate descriptive features of the gait for each stride. 
The following features were extracted: landing step-height, landing step-length, step-clearance, and step-velocity. 
\textit{Landing step-height} ($\mt{h}$) is the vertical position of the forward foot at the end of the stride ($\mt{h} = z_{100}$), this feature changes significantly when transitioning to a new locomotion mode (e.g. from overground walking to stairs) and is positive in the case of ascending and negative in descending strides. 
\textit{Landing step-length} ($\mt{l}$) is the horizontal position of the forward foot at the end of the stride ($\mt{l} = x_{100}$) and represents the distance of the performed stride. 
\textit{Step clearance} ($\mt{c}$) is the maximum vertical position reached from the feet during the stride $\mt{c} = \text{max}_{[0, 100]}(z_k)$. 
\textit{Step-velocity} ($\mt{v}$) is the ratio between the step-length and the time required to execute the step ($\mt{v} = \frac{x_100}{\Delta t}$). 
All of these features were calculated for both the left and right gait phases.
Robot state $\mt{q} = (\theta, \dot{\theta}, \theta_{bp}, \tau_{int}, \alpha, x_{l, r}, z_{l, r})$ and gait features $\mt{f} = (\mt{c}^{gp}_{l, r}, \mt{s}^{gp}_{l, r}, \textbf{h}_{l, r}, \textbf{l}_{l, r}, \textbf{c}_{l, r}, \textbf{v}_{l, r}) \in \mathbb{R}^{12}$ were normalized and sampled to convert from the data logging frequency to 100~Hz. Following these procedures, all data were divided in time windows of 300~ms \cite{s20216345} with a step size of 50ms. 

\subsection{Machine Learning Controller}
\label{sec:machine-learningModel}

In this section, we will present the two ML-based models designed to reduce the complexity of the hierarchical structure. The controller uses a data-driven approach to extract locomotion parameters representing a normative population without additional calibration procedures. Moreover, it allows therapists to modify movement execution in real-time, allowing safer and more structured adaptability to the specific needs of each patient.

Our approach is based on two consecutive neural networks (NN): (1) a Features Extractor Model (FEM), and (2) a Command Predictor Model (CPM). Both models provide probabilistic estimates of the output, enhancing the predictions with information about the associated uncertainty.

\paragraph{Features Extractor Model}

Our FEM (Fig. \ref{fig:concept}.1.A) is based on two bidirectional Long Short-Term Memory (bi-LSTM) networks. The model uses a time window $W$ of robot states ($x_1(t) = [\mt{q}_{t-T_W}, ..., \mt{q}_{t-t_1}, \mt{q}_t]$) as input and returns the current features representation of the step $y_1(t) = \hat{\mt{f}}_t$ as an output.
These features represent the current locomotion intention of the user wearing the exoskeleton; we define them as \textit{self-selected} to distinguish them from those selected by the therapist.

The architecture of the FEM is displayed in Fig. \ref{fig:concept}.A: the model receives a time-window $W$ of dimension 30 of sensor readings and foot positions with a resulting dimension of the input $(30, 15)$. Gaussian noise (standard deviation of 0.01) is used in the first layer to introduce variability to the input data ($x_t$), allowing better generalization. Two bi-LSTM layers allow the model to learn dependencies between the different time steps in the window of data. The output of the second bi-LSTM model is flattened and passed to two consecutive dense layers that transform the features extracted from the bi-LSTMs to a single value representing each feature independently. A \textit{Tanh} activation function is used to bound features in the range of $[-1, 1]$. Each feature is regressed using an independent model to avoid correlation across features. Hyperparameters were selected using a heuristic approach.
The model was trained with the ADAM optimizer, employing the MSE loss function to quantify and minimize the difference between the predicted and actual values. Training proceeded for 100 epochs with early stopping, using batches of 256 samples.

\paragraph{Command Predictor Model}

The second model (Fig. \ref{fig:concept}.1.C) receives as input the ground truth features extracted as described in Sec. \ref{sec:data_collection} ($x_2(t) = \mt{f}_t$) and returns the expected joint kinematics $y_2(t) = (\hat{\mt{\theta}}_t, \hat{\dot{\mt{\theta}}}_t)$ as an output. 
Data augmentation was applied before training by weighted interpolation of both features and kinematics based on the features' Euclidean distances to create a more evenly distributed feature space. For example, we interpolated features from stair ascending and overground walking to generate intermediate feature values, such as moderate step heights.

The structure of the Command Predictor Model is displayed in Fig. \ref{fig:concept}.C. The model receives input of dimensionality $(1,12)$ containing the current features and applies Gaussian noise (standard deviation of 0.01) in the first layer to introduce variability. Then, four consecutive dense layers with reduced dimensions, dropout, and kernel regularization transform the features to a single value representing either the joint position or velocity. Also, in this case, a \textit{Tanh} activation function is used to bound the output in the range of $[-1, 1]$ and then rescaled in the same domain of the kinematics values. Each feature is regressed using an independent model to avoid correlation across features. Hyperparameters were selected using a heuristic approach.
A training procedure similar to those of the Feature extraction models was applied. 

\paragraph{Probabilistic estimation}
\label{sec:prob_estimation}

To have a probabilistic estimation in both layers of the process, six independent NN were trained in parallel using six randomly selected subsets of users. This allows us to estimate and utilize the uncertainty of the predictions. For instance, in the early phases of the swing movement, there will be more uncertainty about the resulting foot placement, resulting in uncertainty in the joint kinematics. With a probabilistic estimation of the features followed by the kinematics, we can model this uncertainty. To propagate the uncertainty across the two phases, we used a Monte-Carlo approach similar to \cite{vianello2021human}: the normal distribution is used to randomly sample $N$ values from the distribution which are passed to the model as a unique batch; the batch of output was fitted to a normal distribution and propagated to the next step of the framework.    

All models presented were trained using TensorFlow (v2.15.0, Google) and converted to TensorFlow Lite (v2.13.0, Google) for real-time inference.

\subsection{Real-time inference and validation}
\label{sec:real-timeInference}

\paragraph{Real-time inference}
The structure of the real-time inference is displayed in Fig. \ref{fig:concept}.
A Python script performing these inferences runs in parallel to the exoskeleton controller running in C++; the two programs communicate through ROS. 
The script (python3 \texttt{concurrent.futures}) calls the multiple FEM Lite models in parallel. The outcome of this first phase (Fig. \ref{fig:concept}.1.A) is a batch of inferences of $\hat{\mt{f}}_i | i \in (0,6)$. The mean and the standard deviation of the extracted features are calculated by fitting the prediction using a normal distribution ($\hat{\mt{f}} \sim \mathcal{N}(\mu_f,\,\sigma_f^{2})\,$).
At this point (Fig. \ref{fig:concept}.1.B), the operator (or therapist) can choose to manually change one of the features using a simplified interface designed using ROS \texttt{rqt}. This choice is implemented as the self-selected mean of the features ($\mu_f$). 
Another parameter that can be manually adjusted is an increment $\delta$, which is added to the mean of the gait phase. This adjustment results in an anticipation of the user's movement by $\delta$ instants, enabling the exoskeleton to deliver proactive assistance\footnote{It is important to note that $\delta$ is a-dimensional because it represents an increment of the gait phase and additional attention is needed to keep the resulting gait phase in the range [0, 100].}.

The distribution $\hat{\mt{f}} \sim \mathcal{N}(\mu_f,\,\sigma_f^{2})\,$ is used to sample a normally distributed batch of features and passed to the CPM model. Also, in this case, the inference on multiple CPM Lite models is executed in parallel and the multiple outputs are fitted to a second normal distribution representing the expected $\delta$ future joint kinematic and its uncertainty $(\hat{\mt{\theta}}, \hat{\dot{\mt{\theta}}}) \sim \mathcal{N}(\mu_{\mt{\theta}, \dot{\mt{\theta}}},\,\sigma_{\mt{\theta}, \dot{\mt{\theta}}}^{2})\,$.  
These values are passed as references to the exoskeleton controller (Eq. \ref{eq:impedanceController}).
In particular, the desired angle and velocity were chosen as equal to the mean of the prediction $\mt{\theta}_d, \dot{\mt{\theta}}_d = \mu_{\mt{\theta}}, \mu_{\dot{\mt{\theta}}}$, while stiffness and damping parameters were selected proportional to the inverse of the standard deviation similarly to~\cite{franzese2021ilosa}:

\begin{align} 
\label{eq:stiff_damp}
    \mt{K} = \mt{K}_S(1 - \frac{\sigma_{\mt{\theta}}}{\sigma_{\text{max}}}) \quad \text{ with } \frac{\sigma_{{\mt{\theta}}}}{\sigma_{\text{max}}} >0, \\
    \mt{D} = \mt{D}_S(1- \frac{\sigma_{\dot{\mt{\theta}}}}{\sigma_{\text{max}}}) \quad \text{ with } \frac{\sigma_{\dot{\mt{\theta}}}}{\sigma_{\text{max}}} >0
\end{align}
where $\sigma_{\text{max}}$ was selected as the maximal uncertainty across the validation set while we assigned  $K_S=[80, 60] \text{ Nm/rad}, D_S = [8, 6]\text{ Nm$\cdot$s/rad}$ for the hip and knee joints. These values were selected in line with previous work identifying impedance parameters during walking \cite{Ortlieb2018Active, Huang2022}.
The full inference process takes approximately $20$ms, corresponding to a frequency of about $50$Hz. The exoskeleton controller updates its reference whenever a new inference is available, while continuously streaming sensor readings at a frequency of approximately $333$Hz.

\paragraph{Validation}

The proposed framework was validated on two healthy participants external to the dataset used to train the two models. 
Participants gave informed consent for their participation. The study protocol was conducted in accordance with the Declaration of Helsinki and approved by the institutional review board of Northwestern University (STU00212684).
The two users performed 15 minutes of treadmill walking and five iterations of stairs ascending and descending. We tested various parameters (i.e., changes in operator-selected features, changes in treadmill speed) to validate the proposed approach by manually changing these parameters during the 15 minutes of walking. 

\section{Results and Discussion}

\begin{figure*}[t]
    \centering
    \vspace{0.2cm}
    \includegraphics[width=0.99\linewidth]{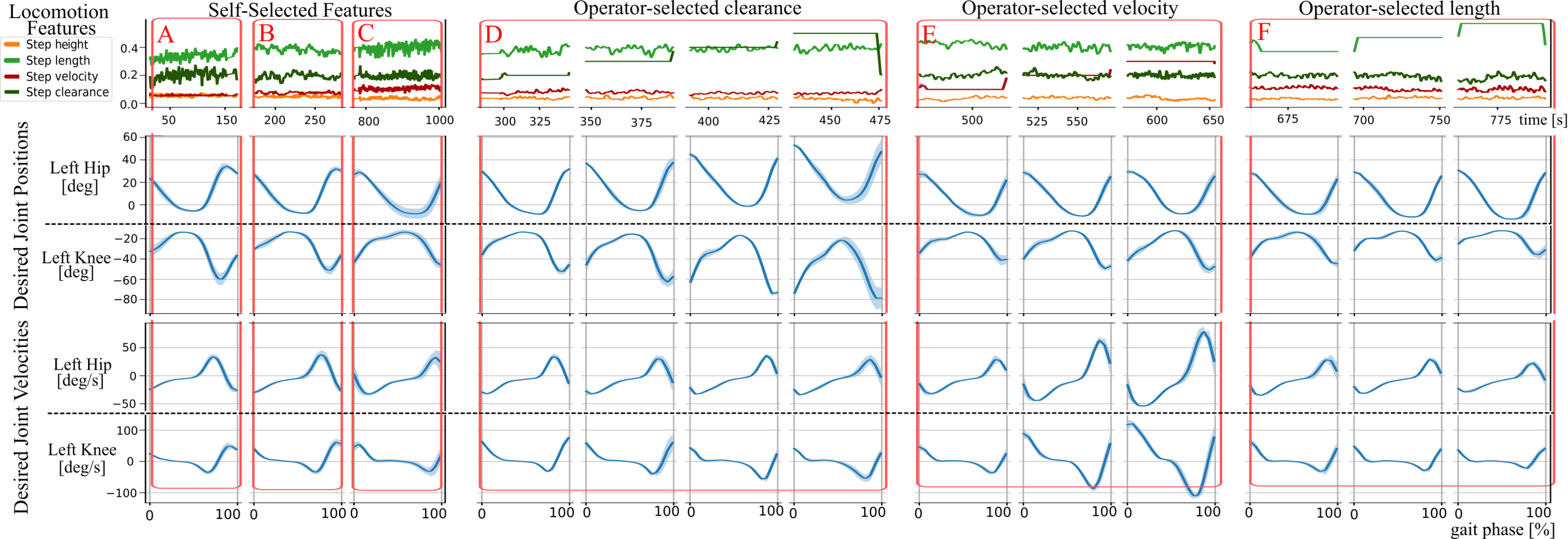}
    \caption{
     Example of walking pattern for a single subject. In order, the five rows show: the implemented features (self-selected or operator-selected) over time [s], the reference position for the left hip and left knee, and the reference velocities for left hip, and left knee over the left stride percentage [\%]. During the trial the user performed: (A) walking with exoskeleton controlled in transparency (self-selected features are not used to close the control loop) with treadmill velocity equal to 0.14 m/s; (B) Self-selected features used to close the loop of the controller; (C) walking with self-selected features on the treadmill at 0.19 m/s; (D) Operator-selected step clearance; (E) Operator-selected step velocity (at the same time increasing the treadmill velocity to 0.19 m/s); (F) Operator-selected step length.}
    \label{fig:results-Hamid}
\end{figure*}

\begin{figure}[t]
    \centering
    \vspace{0.2cm}
    \includegraphics[width=0.95\linewidth]{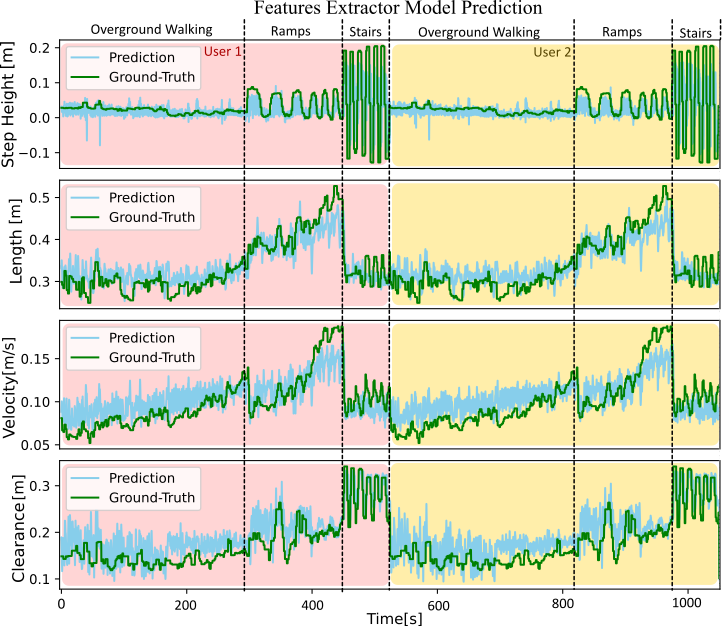}
    \caption{Example of gait feature prediction. The plot displays the ground truth (in green) and the prediction of the Features Extractor Model (in blue) for two users in the training dataset over time (in seconds). Each user (user 1 in pink, user 2 in yellow) performed in order overground walking, ramps (both ascending and descending), and stairs(both ascending and descending). In order top-down the following features are displayed: step height, step length, step velocity, and step clearance. }
    \label{fig:feats_prediction}
\end{figure}

In this section, we present and discuss the results collected in this study. First, we evaluate the offline performance of the two models (FEM and CPM). Next, we present online results demonstrating the proposed controller's ability to assist user locomotion across different locomotion patterns. The results are evaluated using joint kinematics and interaction power metrics for two users.

Fig. \ref{fig:feats_prediction} displays the prediction of gait features for two of the nine users used to develop the FEM and CPM model (seven users for training, two users for validation). The mean absolute error for the features is around a few centimeters (step height $3.1\pm 5.0$cm, step length $2.6 \pm 2.0$cm, step velocity $1.6 \pm 1.1$cm/s, step clearance $2.7 \pm 1.9$cm). Nonetheless, as shown in Fig. \ref{fig:feats_prediction}, the prediction exhibits oscillations caused by uncertainty during the early-swing phase.  Moreover, a limitation of the current approach is the need to identify a set of gait features that adequately capture the complexity of the walking pattern while maintaining clinical relevance. Walking patterns are highly subjective and can vary even for the same user, as shown by a moderate regression accuracy (R$^2$=0.68) between the predicted and desired joint kinematics on the validation set using the CPM. In preliminary tests, we included additional user characteristics (e.g., body weight and height) and features derived using auto-regressive techniques. However, these modifications did not result in additional variance explained (R$^2$=0.68). To address this, we adopted a probabilistic approach (\ref{sec:prob_estimation}) for real-time prediction, which models the uncertainty in gait phases where variability is more prominent. This uncertainty was then used to adapt the compliance of the exoskeleton controller. Future research should focus on integrating additional locomotion features to further enhance the model's predictive capabilities.

\begin{figure*}[t]
    \centering
    \vspace{0.2cm}
\includegraphics[width=0.99\linewidth]{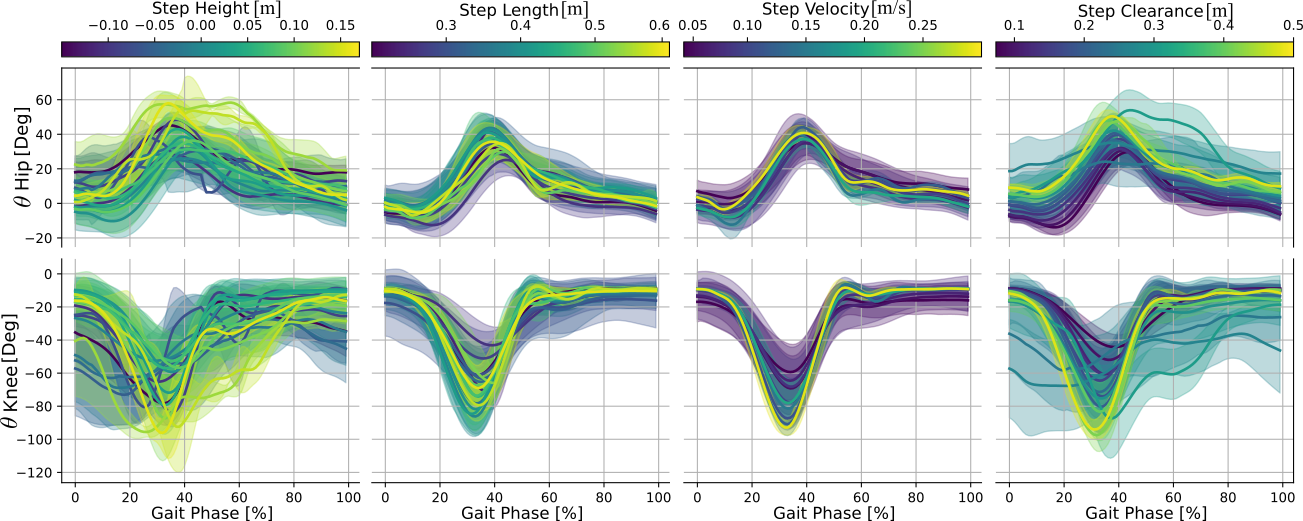}
    \caption{Joint kinematic across gait features (both self-selected and operator-selected): The two rows display respectively the hip and knee joint angles (left and right grouped). Each column represents with a different color code the distribution of the joint kinematic for each gait feature (step-height, step-length, step-velocity, step-clearance). On the top of each column are displayed the color bars for each gait feature. The trend was grouped based on similarities in gait features and is presented using the mean and standard deviation.}
    \label{fig:jointPos}
\end{figure*}

\begin{figure*}[t]
    \centering
    \includegraphics[width=0.99\linewidth]{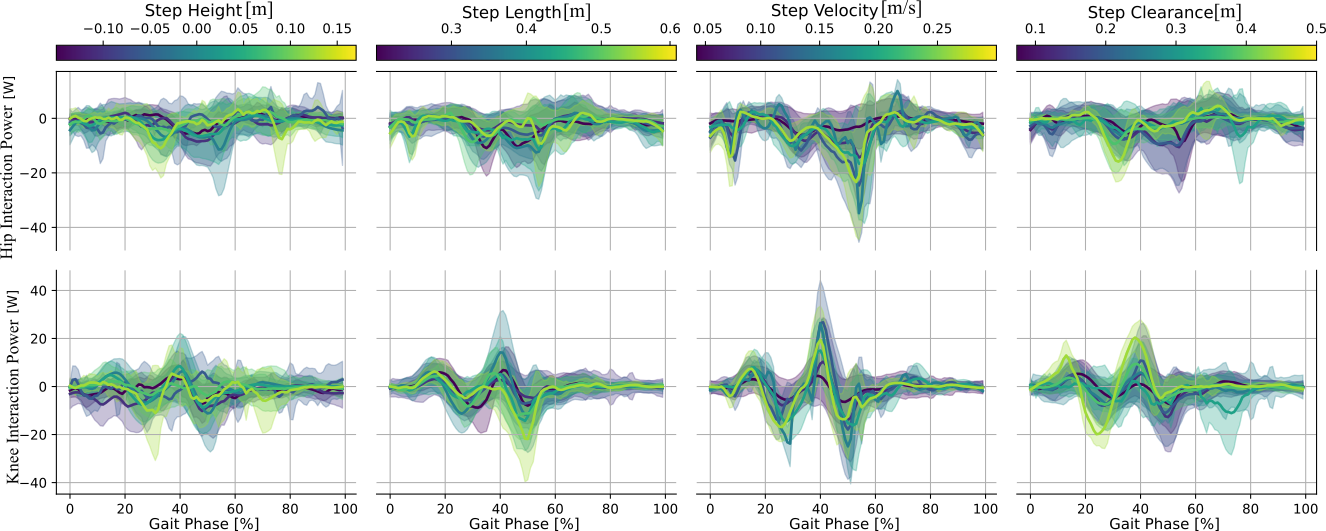}
    \caption{Joint power across gait features (both self-selected and operator-selected): The two rows display the hip and knee joint power (left and right grouped). Each column display with a different color code the distribution of the desired joint kinematic for each gait feature (step-height, step-length, step-velocity, step-clearance).On the top of each column are displayed the color bars for each gait feature. The trend was grouped based on similarities in gait features and is presented using the mean and standard deviation. Negative joint power means exoskeleton assistance to the user while positive joint power represents resistance to the user.}
    \label{fig:intPower}
\end{figure*}

Fig. \ref{fig:results-Hamid} displays an example of a healthy user performing 15 minutes (for a total of 300 strides) of treadmill walking in different conditions. 
During the first three minutes (Fig.\ref{fig:results-Hamid}.A), the user walked at a constant velocity of 0.14 m/s with their self-selected gait pattern while the exoskeleton was controlled in transparency. This first period of walking allows us to verify that the FEM can predict the gait features in an open loop. After predicting these self-selected gait features, we provide closed-loop control by using these features as input to the CPM (Fig.\ref{fig:results-Hamid}.B). The user then walked for two minutes with exoskeleton assistance provided towards their self-selected features. We can observe that closing the loop of the controller (passing from A to B) resulted in an increased step-length (row 1 light-green plots) and step-velocity (row 1 red plot). This is likely due to the assistance the user receives during the movement.
A similar trend can be observed in the following three minutes in which the velocity of the treadmill was increased to 0.19 m/s to validate its effect on the self selected features (Fig.\ref{fig:results-Hamid}.C).

After testing the self-selected features, the operator manually selected the features to evaluate their effects on the participants' walking patterns. Shown in Fig.\ref{fig:results-Hamid}.D, we observe how increasing the step-clearance results in a larger range of motion (ROM) in all kinematic trajectories ($\theta_d$) for both the hip (second row) and the knee (third row). In a similar way, in Fig. \ref{fig:results-Hamid}.E, we observe the impact of increasing the step-speed, corresponding to an increase in the joint velocities ($\dot{\theta}_d$) both for the hip (fourth row) and the knee (fifth row). To account for the increased speed of the user, the velocity of the treadmill was increased from 0.14 m/s to 0.19 m/s and kept constant until the end of the experiment. Finally, in Fig. \ref{fig:results-Hamid}.F, the step length was increased. This resulted in an increased ROM for the hip joint angle ($\theta_d$) and a smaller ROM for the knee angle, likely due to smaller knee flexion during swing movement to increase step-length. 

Fig.\ref{fig:jointPos} shows the distribution of the joint configurations (hip and knee joint) of the two users across the different gait features. We observed that all features impact both the spatial and temporal aspects of the walking pattern. For instance, the step length highly influences the ROM (an increase to a step length of 0.34~m results in a increase of ROM of $20^{\circ}$ for the hip and $40^{\circ}$ for the knee). During these preliminary experiments, we observed an under-damped behavior of the exoskeleton during the early-stance phase (around 60\% of the gait phase). This behavior could be due to two reasons: the impact of training data of stairs (after heel strike during stairs, the knee is more extended than in level-ground walking) or the independence in prediction across joint positions and velocities resulting in uncorrelated stiffness and damping parameters as they were calculated independently (Eq. \ref{eq:stiff_damp}). Further investigation is required to address these limitations.

Fig. \ref{fig:intPower} displays the interaction power ($\tau_{int} \cdot \dot{\theta}$), calculated as a product of the interaction torque and joint velocities. This metric represents the amount of assistance (negative values) or resistance (positive values) that the exoskeleton applied to the user. This metric is displayed for the two joints (hip and knee, grouped) across the two rows, and varying the gait features (across columns). For the hip joint, the interaction power is negative (assistive torque) in most conditions (mean interaction power ($\pm$ std) in all gait phases -2.05 $\pm$ 1.6W. For the knee, the joint interaction power is also negative (-0.6 $\pm$ 1.39W) in most of the gait phase with two peaks of negative interaction power during the swing phase in knee flection (around 30\% of the gait phase) and during both hip and knee extension (around 50-55\% of the gait phase). This suggests that, during swing, the robot is actively assisting the user to execute the movement. On the other hand, two positive peaks were observed at the beginning of the early-swing and the beginning of the late-swing (respectively around 15\% and 40\% of the gait cycle). This can be due to several factors, but most importantly the users were blinded to the gait features manually selected by the operator and thus the nature of the assistance provided. Consequently, they were not aware of how much they should flex the knee during the swing phase. Moreover, they performed each activity for a few minutes, resulting in a limited time to adapt to each condition. Therefore, longer validation sessions are likely required to prove the efficacy of minimizing the interaction power across users.

\section{Conclusions}

In this work, we present a three-step data-driven approach designed to replace the hierarchical structure of traditional exoskeleton controllers. The first step involves probabilistically mapping a short-term history of sensor data (including encoders, force sensors, and foot-plate sensors) to a minimal set of clinically relevant gait features: step length, height (to differentiate ascending and descending surfaces), walking velocity, step clearance, and gait phase. These features can either be applied directly or adjusted by an operator/therapist via a user interface to customize the user's locomotion. The final step uses probabilistic regression to map the current locomotion features to the desired joint posture of the exoskeleton, leveraging prediction uncertainty to adapt the impedance of a spring-damper system linking the user and the device. This adaptive compliance enhances safety, limiting large desired interaction torques when the prediction of the user's behavior is less accurate. Validation on two healthy users across diverse gait activities demonstrated the ability of the proposed approach to detect user intent and provide assistance accordingly, generating predominantly negative interaction power between the user and the exoskeleton.
Future work will explore the selection of additional clinically relevant features in collaboration with therapists to ensure that they align with practical rehabilitation needs. Additionally, more extensive studies involving patient populations (e.g., stroke, spinal cord injury) are necessary to evaluate the impact of these assistive strategies on individuals with gait impairments.

\section*{Acknowledgment}
This work was supported by the National Science Foundation~/~National Robotics Initiative (Grant No: 2024488).
We would like to thank Tim Haswell for his technical support on the hardware improvements of the ExoMotus-X2 exoskeleton.

\bibliographystyle{IEEEtran}
\bibliography{reference, references_SM}

\end{document}